\documentclass{article}
\usepackage{spconf,amsmath,graphicx}
\usepackage{times}
\usepackage{url}
\usepackage{latexsym}
\usepackage{paralist}
\usepackage{multirow}
\usepackage{amssymb,verbatim}

\title{Investigating Linguistic Pattern Ordering \\in Hierarchical Natural Language Generation}
\name{Shang-Yu Su and Yun-Nung Chen}
\address{
National Taiwan University, Taipei, Taiwan\\
\normalsize\texttt{f05921117@ntu.edu.tw \quad y.v.chen@ieee.org}}
\begin{document}
%
\maketitle
\begin{abstract}
Natural language generation (NLG) is a critical component in spoken dialogue system, which can be divided into two phases: (1) sentence planning: deciding the overall sentence structure, (2) surface realization: determining specific word forms and flattening the sentence structure into a string. 
With the rise of deep learning, most modern NLG models are based on a sequence-to-sequence (seq2seq) model, which basically contains an encoder-decoder structure; these NLG models generate sentences from scratch by jointly optimizing sentence planning and surface realization.
However, such simple encoder-decoder architecture usually fail to generate complex and long sentences, because the decoder has difficulty learning all grammar and diction knowledge well.
This paper introduces an NLG model with a hierarchical attentional decoder, where the hierarchy focuses on leveraging linguistic knowledge in a specific order. The experiments show that the proposed method significantly outperforms the traditional seq2seq model with a smaller model size, and the design of the hierarchical attentional decoder can be applied to various NLG systems.
Furthermore, different generation strategies based on linguistic patterns are investigated and analyzed in order to guide future NLG research work\footnote{The source code is available at \url{https://github.com/MiuLab/HNLG}.}.
\end{abstract}
\section{Introduction}
\label{sec:intro}

Spoken dialogue systems that can help users to solve complex tasks have become an emerging research topic in artificial intelligence and natural language processing areas~\cite{wen2017network,li2017end,dhingra2017towards,bordes2017learning}. 
With a well-designed dialogue system as an intelligent personal assistant, people can accomplish certain tasks more easily via natural language interactions. 
Today, there are several virtual intelligent assistants, such as Apple's Siri, Google's Home, Microsoft's Cortana, and Amazon's Alexa, in the market. 
A typical dialogue system pipeline can be divided into several parts: a recognized result of a user's speech input is fed into a natural language understanding module (NLU) to classify the domain along with domain-specific intents and fill in a set of slots to form a semantic frame~\cite{chen2017dynamic, chen2017speaker, su2018how}. 
A dialogue state tracking (DST) module predicts the current state of the dialogue by means of the semantic frames extracted from multi-turn conversations.
Then the dialogue policy determines the system action for the next step given the current dialogue state. 
Finally the semantic frame of the system action is then fed into a natural language generation (NLG) module to construct a response utterance to the user~\cite{wen2015semantically,su2018natural}.

As a key component to a dialogue system, the goal of NLG is to generate  natural language sentences given the semantics provided by the dialogue manager to feedback to users.
As the endpoint of interacting with users, the quality of generated sentences is crucial for better user experience. 
The common and mostly adopted method is the rule-based (or template-based) method~\cite{mirkovic2011dialogue}, which can ensure the natural language quality and fluency.
In spite of robustness and adequacy of the rule-based methods, frequent repetition of identical, tedious output makes talking to a template-based machine unsatisfactory.
Furthermore, scalability is an issue, because designing sophisticated rules for a specific domain is time-consuming~\cite{young2013pomdp}.

Recurrent neural network-based language model (RNNLM) have demonstrated the capability of modeling long-term dependency in sequence prediction by leveraging recurrent structures~\cite{mikolov2010recurrent,mikolov2011extensions}.
Previous work proposed an RNNLM-based NLG that can be trained on any corpus of dialogue act-utterance pairs without hand-crafted features and any semantic alignment~\cite{wen2015stochastic}. 
The following work based on sequence-to-sequence (seq2seq) further obtained better performance by employing encoder-decoder structure with linguistic knowledge such as syntax trees~\cite{cho2014learning,sutskever2014sequence,duvsek2016sequence,bahdanau2014neural}.
However, due to grammar complexity and lack of diction knowledge, it is still challenging to generate long and complex sentences by a simple encoder-decoder structure.

\begin{figure*}[ht]
\centering
\includegraphics[width=0.9\linewidth]{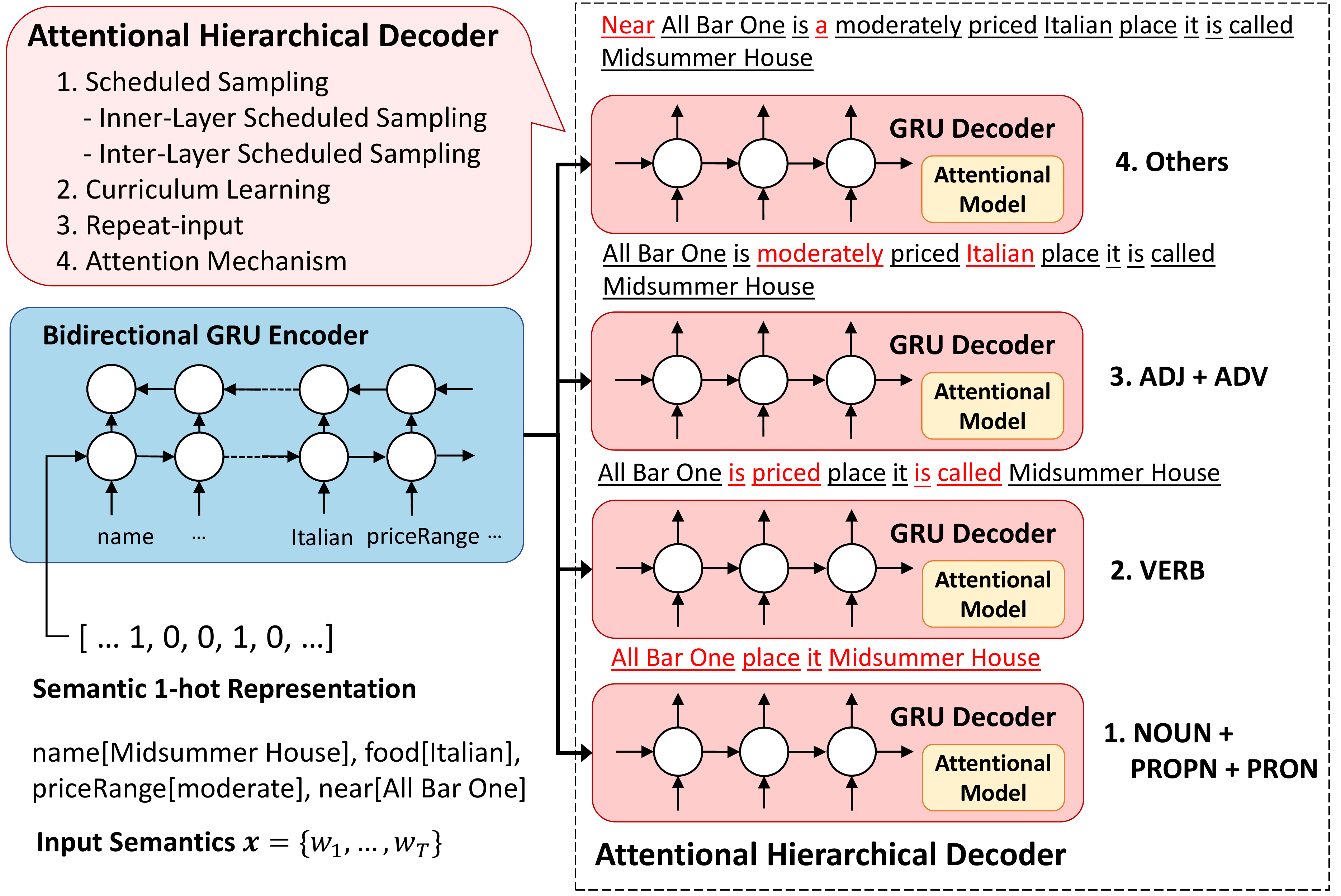}
\vspace{-2mm}
\caption{The illustration of the proposed semantically conditioned NLG model. The hierarchical decoder contains four decoder layer, each is only responsible for learning to insert words of a specific set of POS tags into the sequence.}
\label{fig:framework}
\vspace{-2mm}
\end{figure*}

To address the issue, previous work attempted separating decoding jobs in a decoding hierarchy, which is constructed in terms of part-of-speech (POS) tags~\cite{su2018natural}. 
The original single decoding process is separated into a multi-level decoding hierarchy, where each decoding layer generates words associated with a specific POS set.
This paper extends the idea to a more flexible design by incorporating attention mechanisms into the decoding hierarchy.
Because prior work designs the decoding hierarchy in a hand-crafted manner based on a subjective intuition~\cite{su2018natural}, in this work, we experiment on various generating hierarchies to investigate the importance of linguistic pattern ordering in hierarchical language generation.
The experiments show that our proposed method outperforms the classic seq2seq model with a smaller model size; in addition, the concept of the hierarchical decoder is proven general enough for various generating hierarchies. 
Furthermore, this paper also provides the design guidelines and insights of designing the decoding hierarchy.

\section{Hierarchical Natural Language Generation (HNLG)}
\label{sec:background}

The framework of the proposed hierarchical NLG model is illustrated in Figure~\ref{fig:framework}, where the model architecture is based on an encoder-decoder (seq2seq) structure with attentional hierarchical decoders~\cite{cho2014learning,sutskever2014sequence}.
In the encoder-decoder architecture, a typical generation process includes encoding and decoding phases: 
First, a given semantic representation sequence $\textbf{x}=\{w_t\}^T_1$ is fed into a RNN-based encoder to capture the temporal dependency and project the input to a latent feature space; the semantic representation sequence is also encoded into an one-hot representation as the initial state of the encoder in order to maintain the temporal-independent condition as shown in the left part of Figure~\ref{fig:framework}.
The recurrent unit of the encoder is bidirectional gated recurrent unit (GRU) ~\cite{cho2014learning}, 
\begin{equation}
\label{eq:basic}
\textbf{h}_\text{enc} = \text{BiGRU}(\textbf{x}).
\end{equation}
Then the encoded semantic vector, $\textbf{h}_\text{enc}$, is fed into an RNN-based decoder as the initial state to decode word sequences, as shown in the right part of Figure~\ref{fig:framework}.


\subsection{Attentional Hierarchical Decoder}
\label{ssec:Ahd}

In spite of the intuitive and elegant design of the seq2seq model, it is still difficult to generate complex and decent sequences by a simple encoder-decoder structure, because a single decoder is not capable of learning all diction, grammar, and other related linguistic knowledge at the same time.
Some prior work applied additional techniques such as reranker and beam-search to select a better result among multiple generated sequences~\cite{wen2015stochastic,duvsek2016sequence}.
However, it is still an unsolved issue to the NLG community.

Therefore, we propose a hierarchical decoder to address the above issue, where the core idea is to allow the decoding layers to focus on learning different types of patterns instead of learning all relevant knowledge together.
The hierarchical decoder is composed of several decoding layers, each of which is only responsible for learning a portion of the required knowledge.
Namely, the linguistic knowledge can be incorporated into the decoding process and divided into several subsets.

We use part-of-speech (POS) tags as the additional linguistic features to construct the decoding hierarchy in this paper, where POS tags of the words in the target sentence are separated into several subsets, and each layer is responsible for decoding the words associated with a specific set of POS patterns.
An example is shown in the right part of Figure~\ref{fig:framework}, where the first layer at the bottom is in charge of decoding nouns, pronouns, and proper nouns, and the second layer is for verbs, and so on. 
The prior work manually designed the decoding hierarchy by considering the subjective intuition about how children learn to speak~\cite{su2018natural}: infants first learn to say keywords, which are often nouns.
For example, when an infant says ``\emph{Daddy, toilet.}'', it actually means ``\emph{Daddy, I want to go to the toilet.}''. 
Along with the growth of the age, children learn more grammars and vocabulary and then start adding verbs to the sentences, further adding adverbs, and so on. 
However, the hand-crafted linguistic order may not be optimal, so we experiment and analyze the model on various generating linguistic hierarchies to deeply investigate the effect of linguistic pattern ordering.

In the hierarchical decoder, the initial state of each GRU-based decoding layer $i$ is the extracted feature $\textbf{h}_\text{enc}$ from the encoder, and the input at every step is the last predicted token $\textbf{y}^i_{t-1}$ concatenated with the output from the previous layer $\textbf{y}^{i-1}_t$,
\begin{eqnarray}
\label{eq:basic}
\textbf{h}^i_t, \textbf{o}^i_t &=& \text{GRU}^i_\text{dec}(\textbf{y}^i_{t-1}, \textbf{y}^{i-1}_t \mid \textbf{h}_\text{enc}, \textbf{h}^i_{t-1}), \\
\textbf{y}^i_t &=& \texttt{argmax} (\textbf{o}_t),
\end{eqnarray}
where $\textbf{h}^i_t$ is the $t$-th hidden state of the $i$-th GRU decoding layer and $\textbf{y}^i_t$ is the $t$-th outputted word in the $i$-th layer.
We use the cross entropy loss as our training objective for optimization, where the difference between the predicted distribution and target distribution is minimized.
To facilitate training and improve the performance, several strategies including \emph{scheduled sampling}, a \emph{repeat input mechanism}, \emph{curriculum learning}, and an \emph{attention mechanism} are utilized.

\subsection{Scheduled Sampling}
\label{ssec:interlayertf}
Teacher forcing~\cite{williams1989learning} is a strategy for training RNN that uses model output from a prior time step as an input, and it works by using the expected output at the current time step $\hat{\textbf{y}}_t$ as the input at the next time step, rather than the output generated by the network. 
The teacher forcing techniques can also be triggered only with a certain probability, which is known as the scheduled sampling approach~\cite{bengio2015scheduled}. 
We adopt scheduled sampling methods in our experiments.
In the proposed framework, an input of a decoder contains not only the output from the last step but one from the last decoding layer. Therefore, we design two types of scheduled sampling approaches -- inner-layer and inter-layer.
\begin{itemize}
\item \textbf{Inner-layer schedule sampling} is the classic teacher forcing strategy:
\begin{eqnarray}
\label{eq:basic}
\textbf{h}^i_t, \textbf{o}^i_t = \text{GRU}^i_\text{dec}(\hat{\textbf{y}}^i_{t-1}, \textbf{y}^{i-1}_t \mid \textbf{h}_\text{enc}, \textbf{h}^i_{t-1}).
\end{eqnarray}
\item \textbf{Inter-layer schedule sampling} uses the labels instead of the actual output tokens of the last layer:
\begin{eqnarray}
\label{eq:basic}
\textbf{h}^i_t, \textbf{o}^i_t = \text{GRU}^i_\text{dec}(\textbf{y}^i_{t-1}, \hat{\textbf{y}}^{i-1}_t \mid \textbf{h}_\text{enc}, \textbf{h}^i_{t-1}).
\end{eqnarray}
\end{itemize}


\subsection{Curriculum Learning}
\label{ssec:cl}
The proposed hierarchical decoder consists of several decoding layers, the expected output sequences of upper layers are longer than the ones in the lower layers. 
The framework is suitable for applying the curriculum learning~\cite{elman1993learning}, of which core concept is that a curriculum of progressively harder tasks could significantly accelerate a network’s training. 
The training procedure is to train each decoding layer for some epochs from the bottommost layer to the topmost one.

\subsection{Repeat-Input Mechanism}
\label{ssec:rim}
The concept of the hierarchical decoding is to hierarchically generate the sequence, gradually adding words associated with different linguistic patterns.
Therefore, the generated sequences from the decoders become longer as the generating process proceeds to the higher decoding layers, and the sequence generated by a upper layer should contain the words predicted by the lower layers.
To facilitate the behavior, previous work designs a strategy that repeats the outputs from the last layer as inputs until the current decoding layer outputs the same token, so-called the repeat-input mechanism~\cite{su2018natural}.
This approach offers at least two merits: 
(1) Repeating inputs tells the decoder that the repeated tokens are important to encourage the decoder to generate them. 
(2) If the expected output sequence of a layer is much shorter than the one of the next layer, the large difference in length becomes a critical issue of the hierarchical decoder, because the output sequence of a layer will be fed into the next layer. With the repeat-input mechanism, the impact of length difference can be mitigated.


\begin{table*}[!ht]
\centering
\small
\begin{tabular}{ | l  c c c c|}
    \hline
    \multicolumn{1}{|c}{\bf Generating Linguistic Order} & \bf BLEU & \bf ROUGE-1 & \bf ROUGE-2 & \bf ROUGE-L \\
\hline \hline
\multicolumn{5}{|l|}{\bf ('NOUN', 'PROPN', 'PRON') $\rightarrow$ ('VERB') $\rightarrow$ ('ADJ', 'ADV') $\rightarrow$ (others)} \\
Sequence-to-Sequence Model & 28.89 & 40.75 & 12.52 & 32.05 \\
 + Hierarchical Decoder  & 43.12 & 52.99 & 24.60 & 40.38 \\
 + Hierarchical Decoder, Repeat-Input & 42.33 & 52.91 & 24.03 & 40.08 \\
 + Hierarchical Decoder, Curriculum Learning  & 58.38 & 60.42 & 30.65 & 44.61 \\
 + All & \bf 58.70 & \bf 62.39 & \bf 31.64 & \bf \underline{45.43} \\
\hline
\multicolumn{5}{|l|}{\bf ('NOUN', 'PROPN', 'PRON') $\rightarrow$ ('ADJ', 'ADV') $\rightarrow$ ('VERB') $\rightarrow$ (others)}\\
Sequence-to-Sequence Model & 28.32 & 42.77 & 12.81 & 33.10 \\
+ Hierarchical Decoder  & 43.60 & 53.60 & 25.02 & 40.60 \\
+ Hierarchical Decoder, Repeat-Input & 40.90 & 52.27 & 23.49 & 39.81 \\
+ Hierarchical Decoder, Curriculum Learning  & 58.93 & 60.99 & 30.87 & 44.76 \\
+ All & \bf 59.32 & \bf 62.33 & \bf 32.05 & \bf 45.37 \\
\hline
\multicolumn{5}{|l|}{\bf ('VERB') $\rightarrow$ ('NOUN', 'PROPN', 'PRON') $\rightarrow$ ('ADJ', 'ADV') $\rightarrow$ (others)}\\
Sequence-to-Sequence Model & 28.84 & 39.92 & 11.63 & 31.21 \\
+ Hierarchical Decoder  & 36.60 & 49.90 & 21.85 & 37.70 \\
+ Hierarchical Decoder, Repeat-Input & 35.11 & 48.67 & 20.67 & 37.07 \\
+ Hierarchical Decoder, Curriculum Learning  & 49.29 & 59.65 & 27.85 & 42.98 \\
+ All & \bf 50.73 & \bf 60.76 & \bf 28.74 & \bf 43.53 \\
\hline
\multicolumn{5}{|l|}{\bf ('VERB') $\rightarrow$ ('ADJ', 'ADV') $\rightarrow$ ('NOUN', 'PROPN', 'PRON') $\rightarrow$ (others)}\\
 Sequence-to-Sequence Model & 28.61 & 42.56 & 12.95 & 33.12 \\
+ Hierarchical Decoder  & 40.43 & 51.67 & 23.66 & 39.47 \\
+ Hierarchical Decoder, Repeat-Input & 39.14 & 51.09 & 22.50 & 39.22 \\
+ Hierarchical Decoder, Curriculum Learning  & 58.52 & 61.28 & 31.12 & 44.55 \\
+ All & \bf \underline{61.49} & \bf 62.49 & \bf 31.98 & \bf 45.32 \\
\hline
\multicolumn{5}{|l|}{\bf ('NOUN', 'PROPN', 'PRON') $\rightarrow$ (others) $\rightarrow$ ('VERB')  $\rightarrow$ ('ADJ', 'ADV') }\\
Sequence-to-Sequence Model & 27.72 & 38.92 & 11.56 & 30.52 \\
+ Hierarchical Decoder  & 38.69 & 51.55 & 23.36 & 38.97 \\
+ Hierarchical Decoder, Repeat-Input & 38.48 & 51.76 & 22.98 & 39.10 \\
+ Hierarchical Decoder, Curriculum Learning  & 50.96 & 59.94 & 28.88 & 43.30 \\
+ All & \bf 53.11 & \bf 60.69 & \bf 29.57 & \bf 43.80 \\
\hline
\multicolumn{5}{|l|}{\bf ('NOUN', 'PROPN', 'PRON') $\rightarrow$ (others)  $\rightarrow$ ('ADJ', 'ADV') $\rightarrow$ ('VERB')}\\
Sequence-to-Sequence Model & 29.94 & 43.32 & 13.24 & 33.44 \\
+ Hierarchical Decoder  & 41.78 & 52.56 & 24.56 & 39.97 \\
+ Hierarchical Decoder, Repeat-Input & 40.47 & 52.56 & 22.98 & 39.77 \\
+ Hierarchical Decoder, Curriculum Learning  & \bf 60.50 & 62.65 & \bf \underline{32.66} & 45.41 \\
+ All & 59.46 & \bf \underline{63.20} & 32.28 & \bf 45.47 \\
    \hline
  \end{tabular}
  \vspace{-2mm}
\caption{The proposed attentional hierarchical NLG models with various generating linguistic orders.}
\vspace{-2mm}
\label{tab:res}
\end{table*}

\subsection{Attention Mechanism}
\label{ssec:attn}
In order to model the relationship between layers in a generating hierarchy, we further design attention mechanisms for the hierarchical decoder. The proposed attention mechanisms are content-based, which means the weights are determined based on hidden states of neural models:
\begin{equation}
  \alpha^{l}_{i, j} = 
  \begin{cases}
    (\textbf{h}^{l}_{i})^T \cdot \textbf{h}^{l-1}_{j} & \quad \textbf{Dot Product} \\
    (\textbf{h}^{l}_{i})^T W \textbf{h}^{l-1}_{j}  & \quad \textbf{General} \\
    \tanh(W (\textbf{h}^{l}_{i}, \textbf{h}^{l-1}_{j}))  & \quad \textbf{Concatenation} \\
  \end{cases},
\end{equation}
where $\textbf{h}^{l}_{i}$ is the hidden state at the current step, $\textbf{h}^{l-1}_{j}$ are the hidden states from the previous decoder layer, and $W$ is a learned weight matrix. At each decoding step, attention values $\alpha^{l}_{i, j}$ are calculated by these methods and then used to compute the weighted sum as a context vector, which is then concatenated to decoder inputs as additional information.

\subsection{Training}
The objective of the proposed model is to optimize the conditional probability $p(\mathbf{y}\mid \mathbf{x})$, so that the difference between the predicted distribution and the target distribution, $q(\mathbf{\hat{y}}_\text{k} = z\mid \mathbf{x})$, can be minimized:
\begin{equation}
\mathcal{L}=-\sum_{n=1}^{N}\sum_{k=1}^{K}q(\mathbf{\hat{y}}_\text{k} = z\mid \mathbf{x}) \log p(\mathbf{y}_\text{k} = z\mid \mathbf{x}),
\end{equation}
where $n$ is the number of samples and the labels $\mathbf{\hat{y}}$ are the word labels.
Each decoder in the hierarchical NLG is trained based on curriculum learning with the objective.

\begin{table*}[!ht]
\centering
\small
\begin{tabular}{ | l  c c c c|}
    \hline
    \multicolumn{1}{|c}{\bf Generating Linguistic  Order} & \bf BLEU & \bf ROUGE-1 & \bf ROUGE-2 & \bf ROUGE-L \\
\hline \hline
\multicolumn{5}{|l|}{\bf ('NOUN', 'PROPN', 'PRON') $\rightarrow$ ('VERB') $\rightarrow$ ('ADJ', 'ADV') $\rightarrow$ (others)} \\
All & \bf 58.70 & \bf 62.39 & \bf 31.64 & \bf \underline{45.43} \\
All (Dot-Product Attention) & 56.24 & 61.86 & 30.91 & 44.78 \\
All (General Attention)  & 56.80 & 61.12 & 31.25 & 44.78 \\
All (Concatenation Attention) &  56.13 & 60.14 & 30.11 & 44.56 \\
\hline
\multicolumn{5}{|l|}{\bf ('NOUN', 'PROPN', 'PRON') $\rightarrow$ ('ADJ', 'ADV') $\rightarrow$ ('VERB') $\rightarrow$ (others)}\\
All & \bf 59.32 & \bf 62.33 & \bf 32.05 & \bf 45.37 \\
All (Dot-Product Attention) & 58.93 & 62.26 & 31.83 & 45.04 \\
All (General Attention)  & 57.28 & 62.03 & 31.43 & 44.28 \\
All (Concatenation Attention) &  57.15 & 61.66 & 31.05 & 44.79 \\
\hline
\multicolumn{5}{|l|}{\bf ('VERB') $\rightarrow$ ('NOUN', 'PROPN', 'PRON') $\rightarrow$ ('ADJ', 'ADV') $\rightarrow$ (others)}\\
All & 50.73 & \bf 60.76 & \bf 28.74 & 43.53 \\
All (Dot-Product Attention) & 50.63 & 59.53 & 28.44 & 43.46 \\
All (General Attention)  & 48.53 & 59.82 & 27.50 & 42.87 \\
All (Concatenation Attention) & \bf  50.75 & 59.77 & 28.55 & \bf \underline{44.50} \\
\hline
\multicolumn{5}{|l|}{\bf ('VERB') $\rightarrow$ ('ADJ', 'ADV') $\rightarrow$ ('NOUN', 'PROPN', 'PRON') $\rightarrow$ (others)}\\
All & \bf \underline{61.49} & \bf 62.49 & \bf 31.98 & \bf 45.32 \\
All (Dot-Product Attention) & 59.39 & 61.53 & 31.36 & 44.93 \\
All (General Attention)  & 56.52 & 60.22 & 30.30 & 43.64 \\
All (Concatenation Attention) &  59.20 & 61.83 & 31.48 & 44.86 \\
\hline
\multicolumn{5}{|l|}{\bf ('NOUN', 'PROPN', 'PRON') $\rightarrow$ (others) $\rightarrow$ ('VERB')  $\rightarrow$ ('ADJ', 'ADV') }\\
All & \bf 53.11 & \bf 60.69 & 29.57 & 43.80 \\
All (Dot-Product Attention) & 52.74 & 60.34 & 29.38 & \bf 43.97 \\
All (General Attention)  & 52.64 & 60.68 & \bf 29.67 & 43.59 \\
All (Concatenation Attention) &  50.14 & 58.92 & 28.45 & 43.28 \\
\hline
\multicolumn{5}{|l|}{\bf ('NOUN', 'PROPN', 'PRON') $\rightarrow$ (others)  $\rightarrow$ ('ADJ', 'ADV') $\rightarrow$ ('VERB')}\\
+ All & \bf 59.46 & \bf \underline{63.20} & \bf \underline{32.28} & \bf 45.47 \\
+ All (Dot-Product Attention) & 58.31 & 61.92 & 31.85 & 45.14 \\
+ All (General Attention)  & 57.78 & 62.68 & 32.25 & 44.83 \\
+ All (Concatenation Attention) &  59.05 & 62.01 & 31.66 & 45.38 \\
    \hline
  \end{tabular}
  \vspace{-2mm}
\caption{The proposed hierarchical NLG models with various generating linguistic orders .}
\vspace{-2mm}
\label{tab:attn_results}
\end{table*}

\section{Experiments}
\label{sec:exp}
\subsection{Setup}
\label{ssec:expsetup} 
The E2E NLG challenge dataset~\cite{novikova2017e2e}\footnote{\url{http://www.macs.hw.ac.uk/InteractionLab/E2E/}} is utilized in our experiments, which is a crowd-sourced dataset of 50k instances in the restaurant domain. Our models are trained on the official training set and verified on the official testing set. 
As shown in Figure~\ref{fig:framework}, the inputs are semantic frames containing specific slots and corresponding values, and the outputs are the associated natural language utterances with the given semantics.
For example, a semantic frame with the slot-value pairs ``\texttt{name[Bibimbap House], food[English], priceRange[moderate], area [riverside], near [Clare Hall]}''  corresponds to the target sentence ``\emph{Bibimbap House is a moderately priced restaurant who's main cuisine is English food. You will find this local gem near Clare Hall in the Riverside area.}''.

The data preprocessing includes trimming punctuation marks, lemmatization, and turning all words into lowercase. To prepare the labels of each layer within the hierarchical structure of the proposed method, we utilize spaCy toolkit\footnote{\url{https://spacy.io/}} to perform POS tagging for the target word sequences. Some properties such as names of restaurants are delexicalized (for example, replaced with a symbol ``\texttt{RESTAURANT\_NAME}'') to avoid data sparsity.
In our experiments, we perform six different generating linguistic orders, in which each hierarchy is constructed based on different permutations of the POS tag sets: (1) \textbf{nouns}, \textbf{proper nouns}, and \textbf{pronouns} (2)  \textbf{verbs} (3) \textbf{adjectives} and \textbf{adverbs} (4) \textbf{others}.


The probability of activating inter-layer and inner-layer teacher forcing is set to 0.5, the probability of teacher forcing is attenuated every epoch, and the decaying ratio is 0.9.
The models are trained for 20 training epochs without early stop; when curriculum learning is applied, only the first layer is trained during first five epochs, the second decoder layer starts to be trained at the sixth epoch, and so on.
To evaluate the quality of the generated sequences regarding both precision and recall, the evaluation metrics include BLEU and ROUGE (1, 2, L) scores with multiple references~\cite{lin2004rouge}.

\subsection{Results and Analysis}
\label{ssec:resultandanalysis}

\begin{table*}[!ht]
\small
\centering
\begin{tabular}{ | c | c c c c|}
    \hline
    \multicolumn{1}{|c|}{\multirow{2}{*}{\bf Generating Linguistic Order}} & \multicolumn{4}{c|}{\bf Decoder Layer} \\
 & 1 & 2 & 3 & 4\\
\hline \hline
('NOUN', 'PROPN', 'PRON') $\rightarrow$ ('VERB') $\rightarrow$ ('ADJ', 'ADV') $\rightarrow$ (others) & 6.64/7.90 & ~9.67/11.53 & 12.54/14.84 & 18.09/21.32 \\
('NOUN', 'PROPN', 'PRON') $\rightarrow$ ('ADJ', 'ADV') $\rightarrow$ ('VERB') $\rightarrow$ (others) & 6.64/7.90 & ~9.51/11.21 & 12.54/14.84 & 18.09/21.32 \\
('VERB') $\rightarrow$ ('NOUN', 'PROPN', 'PRON') $\rightarrow$ ('ADJ', 'ADV') $\rightarrow$ (others) & 3.03/3.62 & ~9.67/11.53 & 12.54/14.84 & 18.09/21.32 \\
('VERB') $\rightarrow$ ('ADJ', 'ADV') $\rightarrow$ ('NOUN', 'PROPN', 'PRON') $\rightarrow$ (others) & 3.03/3.62 & ~5.91/~~6.94 & 12.54/14.84 & 18.09/21.32 \\
('NOUN', 'PROPN', 'PRON') $\rightarrow$ (others) $\rightarrow$ ('VERB')  $\rightarrow$ ('ADJ', 'ADV')  & 6.64/7.90 & 12.18/14.38 & 15.21/18.01 & 18.09/21.32 \\
('NOUN', 'PROPN', 'PRON') $\rightarrow$ (others)  $\rightarrow$ ('ADJ', 'ADV') $\rightarrow$ ('VERB') & 6.64/7.90 & 12.18/14.38 & 15.06/17.70 & 18.09/21.32 \\
    \hline
  \end{tabular}
  \vspace{-2mm}
\caption{The average length of the target sequences for each decoder layer in the training data (left) and testing data (right).}
\vspace{-2mm}
\label{tab:len}
\end{table*}

In the experiments, we borrow the idea of hierarchical decoding proposed by the previous work~\cite{su2018natural} and investigate various extensions of generating hierarchies.
To examine the effectiveness of hierarchical decoders, we control our model size to be smaller than the baseline's.
Specifically, the decoder in the baseline seq2seq model has hidden layers of size 400, while our models with hierarchical decoders have four decoding layers of size 100 for fair comparison.

\subsubsection{Effectiveness of Hierarchical Decoders}
Table~\ref{tab:res} compares the performance between a baseline and proposed models with different generating linguistic orders.
For all generating hierarchies with different orders, simply replacing the decoder by a hierarchical decoder achieves significant improvement in every evaluation metrics;
for example, the topmost generating hierarchy in Table~\ref{tab:res} has 49.25\% improvement in BLEU, 30.03\% in ROUGE-1, 96.48\% in ROUGE-2, and 25.99\% in ROUGE-L respectively.
In other words, separating the generation process into several phases is proven to be a promising method. Performing curriculum learning strategy offers a considerable improvement, take the topmost generating hierarchy in Table~\ref{tab:res} for example, this method yields a 102.07\% improvement in BLEU, 48.26\% in ROUGE-1, 144.8\% in ROUGE-2, and 39.18\% in ROUGE-L. Despite that applying repeat-input mechanism alone does not offer benefit, combining these two strategies together further achieves the best performance. Note that these methods do not require any additional parameters.

\subsubsection{Effectiveness of Attention Mechanism}
Unfortunately, even some of the attentional hierarchical decoders achieve the best results in the generating hierarchies (Table~\ref{tab:attn_results}). 
Mostly, the additional attention mechanisms are not capable of bringing benefit for model performance. The reason may be that the decoding process is designed for gradually importing words in the specific set of linguistic patterns to the output sequence, each decoder layer is responsible of copying the output tokens from the previous layer and insert new words into the sequence precisely. Because of this nature, a decoder needs explicit information of the structure of a sentence rather than implicit high-level latent information. For instance, when a decoder is trying to insert some Verb words into the output sequence, knowing the position of subject and object would be very helpful.

\subsubsection{Analysis of Linguistic Orders}
The above results show that among these six different generating hierarchy, the generating order: (1) \textbf{verbs} $\rightarrow$ (2) \textbf{nouns}, \textbf{proper nouns}, and \textbf{pronouns}  $\rightarrow$ (3) \textbf{adjectives} and \textbf{adverbs}  $\rightarrow$ (4) the other POS tags  yields the worst performance. 
Table~\ref{tab:len} shows that the gap of average length of target sequences between the first and the second decoder layer is the largest among all the hierarchies; in average, the second decoder needs to insert up to 8 words into the sequence based on 3.62 words from the first decoder layer  in this generation process, which is absolutely difficult. The essence of the hierarchical design is to separate the job of the decoder into several phases; if the job of each phase is balanced, it is intuitive that it is more suitable for applying curriculum learning and improve the model performance.

The model performance is also related to linguistic structures of sentences: the fifth and the sixth generating hierarchies in Table~\ref{tab:res} have very similar trends, where the length of target sentences of each decoder layer is almost identical as shown in Table~\ref{tab:len}.
However, the model performance differs a lot.
An adverb word could be used to modify anything but nouns and pronouns, which means that the number of adverbs used for modifying verbs would be a factor to determine the generating order as well.
In our cases, almost all adverbs in the dataset are used to describe adjectives, indicating that generating verbs before inserting adverbs to sequences may not provide enough useful information; instead, it would possibly obstruct the model learning. 
We can also find that in all experiments, inserting adverbs before verbs would be better.

In summary, the concept of the hierarchical decoder is simple and useful, separating a difficult job to many phases is demonstrated to be a promising direction and not limited to a specific generating hierarchy. 
Furthermore, the generating linguistic orders should be determined based on the dataset, and the important factors include the distribution over length of subsequences and the linguistic nature of the dataset for designing a proper generating hierarchy in NLG.

\section{Conclusion}
\label{sec:conclusion}
This paper investigates the seq2seq-based model with a hierarchical decoder that leverages various linguistic patterns. The experiments on different generating linguistic orders demonstrates the generalization about the proposed hierarchical decoder, which is not limited to a specific generating hierarchy.
However, there is no universal decoding hierarchy, while the main factor for designing a suitable generating order is the nature of the dataset.

\section{Acknowledgements}
We would like to thank reviewers for their insightful comments on the paper. This work was financially supported by Ministry of Science and Technology (MOST) in Taiwan.

\bibliographystyle{IEEEbib}
\bibliography{refs.bib}

\end{document}